\pdfoutput=1

\documentclass[11pt]{article}

\usepackage{acl}

\usepackage{times}
\usepackage{latexsym}

\usepackage[T1]{fontenc}

\usepackage[utf8]{inputenc}

\usepackage{microtype}
\usepackage{multirow}
\usepackage{latexsym}
\usepackage{booktabs}
\usepackage{courier}
\usepackage{amsmath}
\usepackage{amsfonts} 
\usepackage{subfig}
\usepackage{placeins}
\usepackage{tikz}
\usepackage{hyperref}

\usepackage{soul}
\usepackage{color}

\title{Multi-Relational Hyperbolic Word Embeddings \\from Natural Language Definitions}

\author{Marco Valentino$^1$, Danilo S. Carvalho$^{2,3}$, Andr\'e Freitas$^{1,2,3}$ \\
$^{1}$ Idiap Research Institute, Switzerland\\
$^{2}$ Department of Computer Science, University of Manchester, UK\\
$^{3}$ National Biomarker Centre, CRUK-MI, University of Manchester, UK\\
\texttt{$^{1}$\{firstname.lastname\}@idiap.ch}
\\ \texttt{$^{2}$\{firstname.lastname\}@manchester.ac.uk}}

\begin{document}
\maketitle
\begin{abstract}

\emph{Natural language definitions} possess a recursive, self-explanatory semantic structure that can support representation learning methods able to preserve explicit conceptual relations and constraints in the latent space.
This paper presents a \emph{multi-relational} model that explicitly leverages such a structure to derive \emph{word embeddings} from definitions.
By automatically extracting the relations linking defined and defining terms from dictionaries, we demonstrate how the problem of learning word embeddings can be formalised via a translational framework in \emph{Hyperbolic space} and used as a proxy to capture the global semantic structure of definitions. An extensive empirical analysis demonstrates that the framework can help imposing the desired structural constraints while preserving the semantic mapping required for controllable and interpretable traversal. Moreover, the experiments reveal the superiority of the Hyperbolic word embeddings over the Euclidean counterparts and demonstrate that the multi-relational approach can obtain competitive results when compared to state-of-the-art neural models, with the advantage of being intrinsically more efficient and interpretable\footnote{Code and data available at: \url{https://github.com/neuro-symbolic-ai/multi_relational_hyperbolic_word_embeddings}}.

\end{abstract}

\section{Introduction}

A \emph{natural language definition} is a statement whose core function is to describe the essential meaning of a word or a concept. As such, extensive collections of definitions \cite{miller1995wordnet,zesch2008using}, such as the ones found in dictionaries or technical discourse, are often regarded as rich and reliable sources of information from which to derive textual embeddings \cite{tsukagoshi2021defsent,bosc2018auto,tissier2017dict2vec,noraset2017definition,hill2016learning}.

\begin{figure}[t]
\centering
\includegraphics[width=0.95\columnwidth]{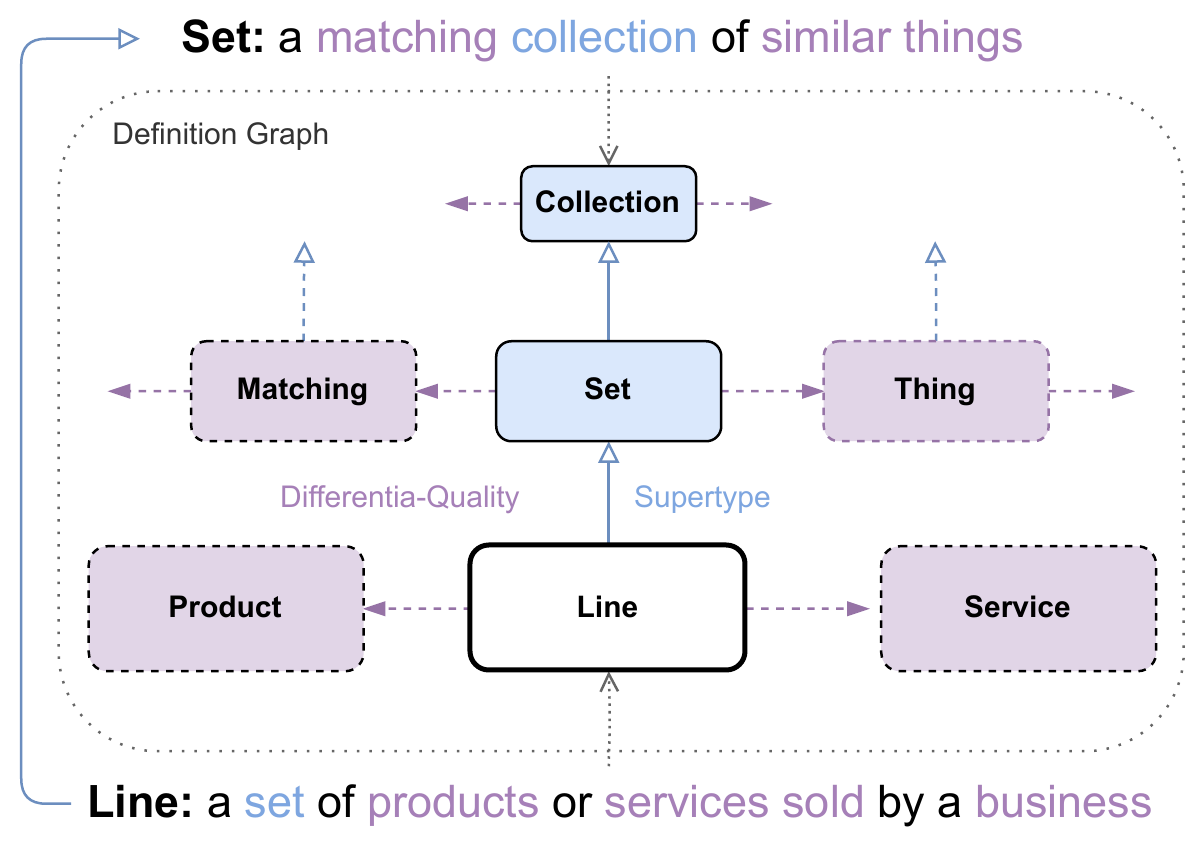}
\caption{How can we inject the recursive, hierarchical structure of \emph{natural language definitions} into word embeddings? This paper investigates \emph{Hyperbolic manifolds} to learn \emph{multi-relational} representations exclusively from definitions, formalising the problem via a translational framework to preserve the semantic mapping between concepts in latent space.}
\label{fig:example}
\end{figure}

A fundamental characteristic of natural language definitions is that they are widely abundant, possessing a \emph{recursive, self-explanatory} semantic structure which typically connects the meaning of terms composing the definition (\emph{definiens}) to the meaning of the terms being defined (\emph{definiendum}). This structure is characterised by a well-defined set of semantic roles linking the terms through explicit relations such as subsumption and differentiation  ~\cite{silva2016categorization} (see Figure \ref{fig:example}). 
However, existing paradigms for extracting embeddings from natural language definitions 
rarely rely on such a structure, often resulting in poor interpretability and semantic control \cite{mikolov2013efficient,pennington2014glove,reimers2019sentence}.

This paper investigates new paradigms to overcome these limitations. Specifically, we posit the following research question: \emph{``How can we leverage and preserve the explicit semantic structure of natural language definitions for neural-based embeddings?''}
To answer the question, we explore \emph{multi-relational} models that can learn to explicitly map \emph{definenda}, \emph{definiens}, and their corresponding \emph{semantic relations} within a continuous vector space. Our aim, in particular, is to build an embedding space that can encode the structural properties of the relevant semantic relations, such as concept hierarchy and differentiation, as a product of geometric constraints and transformations. The multi-relational nature of such embeddings should be intrinsically interpretable, and define the movement within the space in terms of mapped relations and entities. Since \emph{Hyperbolic manifolds} have been demonstrated to correspond to continuous approximations of recursive and hierarchical structures \cite{nickel2017poincare}, we hypothesise them to be the key to achieve such a goal.

Following these motivations and research hypotheses, we present a \emph{multi-relational framework} for learning \emph{word embeddings} exclusively from \emph{natural language definitions}. Our methodology consists of two main phases. First, we build a \emph{specialised semantic role labeller} to automatically extract multi-relational triples connecting definienda and definiens. This explicit mapping allows casting the learning problem into a \emph{link prediction} task, which we formalise via a \emph{translational} objective \cite{balazevic2019multi,feng2016knowledge,bordes2013translating}. By specialising the translational framework in Hyperbolic space through \emph{Poincaré manifolds}, we are able to jointly embed entities and semantic relations, imposing the desired structural constraints while preserving the explicit mapping for a controllable traversal of the space.

An extensive empirical evaluation led to the following conclusions:
\begin{enumerate}
    \item  Instantiating the multi-relational framework in Euclidean and Hyperbolic spaces reveals the explicit gains of Hyperbolic manifolds in capturing the global semantic structure of definitions. The Hyperbolic embeddings, in fact, outperform the Euclidean counterparts on the majority of the benchmarks, being also superior on one-shot generalisation experiments designed to assess the structural organisation and interpretability of the embedding space.
    \item A comparison with distributional approaches and previous work based on autoencoders demonstrates the impact of the semantic relations on the quality of the embeddings. The multi-relational model, in fact, outperforms previous approaches with the same dimensions, while being intrinsically more interpretable and controllable. 
    \item The multi-relational framework is competitive with state-of-the-art Sentence-Transformers, having the advantage of requiring less computational and training resources, and possessing a significantly lower number of dimensions.
    \item We conclude by performing a set of qualitative analyses to visualise the interpretable nature of the traversal for such vector spaces. We found that the multi-relational framework enables robust semantic control, clustering the closely defined terms according to the target semantic transformations. 
\end{enumerate}

To the best of our knowledge, we are the first to conceptualise and instantiate a multi-relational Hyperbolic framework for representation learning from natural language definitions, opening new research directions for improving the interpretability and structural control of neural embeddings. 

\begin{figure*}[t]
\centering
\includegraphics[width=0.935\textwidth]{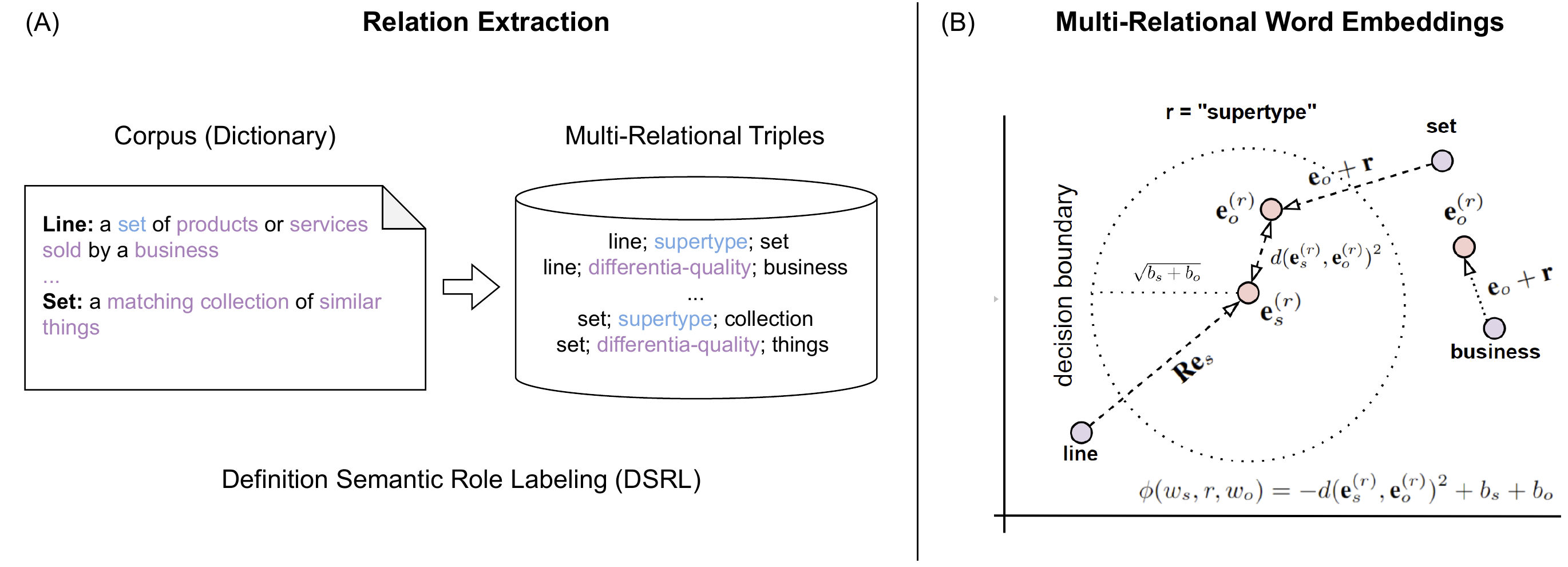}
\caption{An overview of the multi-relational framework for learning word embeddings from definitions. The methodology consists of two main phases: (A) building a specialised \emph{semantic role labeller} (DSRL) for the annotation of natural language definitions and the extraction of relations from large dictionaries; (B) formalising the learning problem as a \emph{link prediction} task via a translational framework. The translational formulation acts as a proxy for minimising the distance between words that are connected in the definitions (e.g., \emph{line} and \emph{set}) while preserving the semantic relations for interpretable and controllable traversal of the space.}
\label{fig:approach}
\end{figure*}

\section{Background}

\subsection{Natural Language Definitions}
\label{sec:dictionary_definitions}

\begin{table}[t]
    \small
    \centering
    \begin{tabular}{p{2.5cm}p{4.5cm}}
        \toprule
        \textbf{Role} & \textbf{Description}\\ 
        \midrule
        \textbf{Supertype} & An hypernym for the definiendum. \\
        \midrule
        \textbf{Differentia Quality} & A quality that distinguishes the definiendum from other concepts under the same supertype. \\
        \midrule
        \textbf{Differentia Event} & an event (action, state or process) in which the definiendum participates and is essential to distinguish it from other concepts under the same supertype. \\
        \midrule
        \textbf{Event Location} & the location (spatial or abstract) of a differentia event. \\
        \midrule
        \textbf{Event Time} & the time in which a differentia event happens. \\
        \midrule
        \textbf{Origin Location} & the definiendum's location of origin. \\
        \midrule
        \textbf{Quality Modifier} & degree, frequency or manner modifiers that constrain a differentia quality. \\
        \midrule
        \textbf{Purpose} & the main goal of the definiendum's existence or occurrence. \\
        \midrule
        \textbf{Associated Fact} & a fact whose occurrence is/was linked to the definiendum's existence or occurrence. \\
        \midrule
        \textbf{Accessory Determiner} & a determiner expression that doesn't constrain the supertype / differentia scope. \\
        \midrule
        \textbf{Accessory Quality} & a quality that is not essential to characterize the definiendum. \\
        \bottomrule
    \end{tabular}
    \caption{The complete set of Definition Semantic Roles (DSRs) considered in this work.}
    \label{tab:dsr_description}
\end{table}

Natural language definitions possess a \emph{recursive, self-explanatory} semantic structure. Such structure connects the meaning of terms composing the definition (\emph{definiens}) to the meaning of the terms being defined (\emph{definiendum}) through a set of semantic roles (see Table \ref{tab:dsr_description}). These roles describe particular semantic relations between the concepts, such as subsumption and differentiation~\cite{silva2016categorization}. Previous work has shown the possibility of automated categorisation of these semantic roles~\cite{silva2018building}, and leveraging those can lead to models with higher interpretability and better navigation control over the semantic space \cite{carvalho2022learning,silva2019exploring,silva2018recognizing}. 

It is important to notice that while the definitions are lexically indexed by their respective definienda, the terms they define are \textit{concepts}, and thus a single lexical item (definiendum) can have multiple definitions. For example, the word ``line'' has the following two definitions, among others:\\

``\textit{An infinitely extending one-dimensional figure that has no curvature.}''

``\textit{A set of products or services sold by a business, or by extension, the business itself.}''\\

from which upon analysis, we can find the roles of \textbf{supertype} and \underline{differentia quality}, as follows:\\

``\textit{An \underline{infinitely extending one-dimensional} \textbf{figure} \underline{that has no curvature}.}''

``\textit{A \textbf{set} \underline{of products or services sold by a business}, or \underline{by extension, the business itself}.}''\\

A definiendum can then be identified by the interpretation of its associated terms, categorised according to its semantic roles within the definition. A line which has ``figure'' as \textit{supertype} is thus a different concept from a line which has ``set'' as \textit{supertype}. The same can be applied for the other aforementioned roles: a line with supertype ``set'' and distinguished by the \textit{differentia quality} ``product'' is different from a line distinguished by ``point'' on the same role. This is a recursive process, as each term in a definition is also representing a concept, which may be defined in the dictionary. This entails a hierarchical and multi-relational structure linking the terms in the definiendum and in the definiens.

\subsection{Hyperbolic Embeddings}

As the semantic roles induce multiple hierarchical and recursive structures (e.g., the \texttt{supertype} and \texttt{differentia quality} relation), we hypothesise that Hyperbolic geometry can play a crucial role in learning word embeddings from definitions. 
Previous work, in fact, have demonstrated that recursive and hierarchical structures such as trees can be represented in a continuous space via a d-dimensional Poincaré ball \cite{nickel2017poincare,balazevic2019multi}. 

A Poincaré ball ($\mathbb{B}^d_c$, $g^\mathbb{B}$) of radius $1/\sqrt{c}, c > 0$ is a
d-dimensional manifold equipped with the Riemannian metric $g^\mathbb{B}$. In such d-dimensional space, the distance between two vectors $\textbf{x},\textbf{y}\in\mathbb{B}$ can be computed along a geodesic as follows:


\begin{equation}
\label{eq:poincare_distance}
\small
d_\mathbb{B}(\textbf{x}, \textbf{y}) = \frac{2}{\sqrt{c}} \tanh^{-1}\left(\sqrt{c}\| - \textbf{x} \oplus_{c} \textbf{y}\|\right),
\end{equation}

\noindent where $\lVert \cdot \rVert$ denotes the Euclidean norm and $\oplus_c$ represents Mobi\"us addition \cite{ungar2001Hyperbolic}:

\begin{equation}
\label{eq:mobius_add}
\small
    \textbf{x} \oplus \textbf{y} = \frac{(1+2c\langle \textbf{x},\textbf{y} \rangle + c\|\textbf{y}\|^2)\textbf{x} + (1-c\|\textbf{x}\|^2)\textbf{y}}{1+2c\langle \textbf{x},\textbf{y} \rangle + c^2\|\textbf{x}\|^2\|\textbf{y}\|^2},
\end{equation}
with $\langle \cdot,\cdot \rangle$ representing the Euclidean inner product. A crucial feature of Equation \ref{eq:poincare_distance} is that it allows determining the organisation of hierarchical structures locally, simultaneously capturing the
hierarchy of entities (via the norms) and their similarity (via the distances) \cite{nickel2017poincare}.

Remarkably, subsequent work has shown that this formalism can be extended for multi-relational graph embeddings via a translation framework \cite{balazevic2019multi}, parametrising multiple Poincaré balls within the same embedding space (Section \ref{sec:multi_relational_embeddings}).

\section{Methodology}

We present a \emph{multi-relational} model to learn \emph{word embeddings} exclusively from \emph{natural language definitions} that can leverage and preserve the \emph{semantic relations} linking definiendum and definiens. 

The methodology consists of two main phases: (1) building a specialised \emph{semantic role labeller} (DSRL) for the automatic annotation of natural language definitions from large dictionaries; (2) formalising the task of learning multi-relational word embeddings as a \emph{link prediction} problem via a translational framework. 


\subsection{Definition Semantic Roles (DSRs)}
\label{sec:definition_semantic_roles}


Given a natural language definition $D = \{w_1, \ldots, w_n\}$ including terms $w_1, \ldots, w_n$ and semantic roles $SR = \{r_1, \ldots, r_m\}$, we aim to build a DSRL that assigns one of the semantic roles in $SR$ to each term in $D$. 
To this end, we explore the fine-tuning of different versions of BERT framing the task as a token classification problem \cite{devlin2019bert}.
To fine-tune the models, we adopt a publicly available dataset of $\approx$ 4000 definitions extracted from Wordnet, each manually annotated with the respective semantic roles\footnote{\url{https://drive.google.com/drive/folders/12nJJHo7ryS6gVT-ukE-BsuHvAqPLUh3S?usp=sharing}} \cite{silva2016categorization}. Specifically, we annotate the definition sentences using BERT to annotate each token with a semantic role (i.e., supertype, differentia-quality, etc.). After the annotation, as we aim to learn word embeddings, we map back the tokens to the original words and use the associated semantic roles to construct multi-relational triples (Section \ref{sec:multi_relational_embeddings}).

Overall, we found \texttt{distilbert} \cite{sanh2019distilbert} to achieve the best trade-off between efficiency and accuracy (86\%), obtaining performance comparable to \texttt{bert-base-uncased} while containing 40\% less parameters. Therefore, we decided to employ \texttt{distilbert} for subsequent experiments. While more accurate DSRLs could be built via the fine-tuning of more recent Transformers, we regard this trade-off as satisfactory for current purposes. 

Table \ref{fig:dsrl_micro_results} reports the detailed results achieved by different versions of BERT in terms of precision, recall, f1 score, and accuracy. To train the models, we adopted a k-fold cross-validation technique with $k = 5$, fine-tuning the models for 3 epochs in total via Huggingface\footnote{\url{https://huggingface.co/}}. 

\begin{table}[t]
    \small
    \centering
    \begin{tabular}{lcccc}
        \toprule
                & \textbf{P} & \textbf{R} & \textbf{F1} & \textbf{Acc.} \\ 
        \midrule
        \texttt{bert-base-uncased} & 0.76 & 0.80 & 0.78 & 0.86 \\
        \texttt{bert-large-uncased} & 0.69 & 0.77 & 0.73 & 0.85\\ \midrule
        \texttt{distilbert}  & 0.76 & 0.79 &  0.77 & 0.86 \\\bottomrule
    \end{tabular}
    \caption{Micro-average results for the Definition Semantic Role Labeling (DSRL) task using different versions of BERT \cite{devlin2019bert}.}
    \label{fig:dsrl_micro_results}
\end{table}

\subsection{Multi-Relational Word Embeddings}
\label{sec:multi_relational_embeddings}
Thanks to the semantic annotation, it is possible to leverage the relational structure of natural language definitions for training word embeddings. Specifically, we rely on the semantic roles to cast the task into a \emph{link prediction} problem. Given a set of definiendum-definition pairs, we first employ the DSRL to automatically annotate the definitions, and subsequently extract a set of subject-relation-object triples of the form $(w_s, r, w_o)$, where $w_s$ represents a defined term, $r$ a semantic role, and $w_o$ a term appearing in the definition of $w_s$ with semantic role $r$. To derive the final set of triples for training, we remove the instances in which $w_o$ represents a stop-word.  

In order to train the word embeddings, the link prediction problem is formalised via a \emph{translational objective function} $\phi(\cdot)$:

\begin{equation}
\label{eq:general_translational_objective}
    \small
    \begin{aligned}
        \phi(w_s, r,w_o) = -d(\textbf{e}_s^{(r)}, \textbf{e}_o^{(r)})^2 + b_s + b_o \\ = -d(\textbf{R}\textbf{e}_s, \textbf{e}_o + \textbf{r})^2 + b_s + b_o,
    \end{aligned}
\end{equation}
\noindent where $d(\cdot)$ is a generic distance function, \textbf{e}$_s$,\textbf{e}$_o \in \mathbb{R}^d$ represent the embeddings of $w_s$ and $w_o$ respectively, and $b_s$, $b_o\in \mathbb{R}$ act as scalar biases for subject and object word. On the other hand, \textbf{r} $\in\mathbb{R}^d$ is a translation vector encoding the semantic role $r$, while \textbf{R} $\in \mathbb{R}^{d\times d}$ is a diagonal relation matrix. Therefore, the output of the objective function $\phi(w_s, r,w_o)$ is directly proportional to the similarity between $\textbf{e}_s^{(r)}$ and $\textbf{e}_o^{(r)}$, which represent the subject and object word embedding after applying a relation-adjusted transformation.

The choice behind the translational formulation is dictated by a set of goals and research hypotheses. First, we hypothesise that the global multi-relational structure of dictionary definitions can be optimised locally via the extracted semantic relations (i.e., making words that are semantically connected in the definitions closer in the latent space). Second, the translational formulation allows for the joint embedding of words and semantic roles. This plays a crucial function as it enables the explicit parametrisation of multiple relational structures within the same vector space (i.e., with each semantic role vector acting as a geometrical transformation), and second, it allows for the explicit use of the semantic roles after training. By preserving the embeddings of the semantic relations, in fact, we aim to make the vector space intrinsically more interpretable and controllable.

\paragraph{Hyperbolic Model.} 

Following previous work on multi-relational Poincaré embeddings \cite{balazevic2019multi}, we specialise the general translational objective function in Hyperbolic space:

\begin{equation}
    \small
    \begin{aligned}
        \phi_\mathbb{B}(w_s,r,w_o) = -d_\mathbb{B}(\textbf{h}_s^{(r)}, \textbf{h}_o^{(r)})^2 + b_s + b_o \\ = -d_\mathbb{B}(\textbf{R}\otimes_c\textbf{h}_s, \textbf{h}_o \oplus_c \textbf{r})^2 + b_s + b_o,
    \end{aligned}
\end{equation}
\noindent where $d_\mathbb{B}(\cdot)$ is the Poincaré distance, $\textbf{h}_s$, $\textbf{h}_o$, \textbf{r} $\in \mathbb{B}^{d}_c$ are the hyperbolic embeddings of words and semantic roles, $\textbf{R} \in \mathbb{R}^{d\times d}$ is a diagonal relation matrix, $\oplus$ and $\otimes$ represents Mobi\"us addition (Equation \ref{eq:mobius_add}) and matrix-vector multiplication \cite{ganea2018Hyperbolic}:

\begin{equation}
\small
    \textbf{R} \otimes_c \textbf{h} = \exp^c_0 (\textbf{R} \log^c_0 (\textbf{h})),
\end{equation}
with $\log(\cdot)$ and $\exp(\cdot)$ representing the logarithmic and exponential maps for projecting a point into the Euclidean tangent space and back to the Poincaré ball.

\paragraph{Training \& Optimization.} The multi-relational model is optimised for link prediction via the Bernoulli negative log-likelihood loss (details in Appendix \ref{sec:AppendixA}).
We employ Riemmanian optimization to train the Hyperbolic embeddings, enriching the set of extracted triples with random negative sampling. We found that the best results are obtained with 50 negative examples for each positive instance. In line with previous work on Hyperbolic embeddings \cite{balazevic2019multi} we set $c = 1$. Following guidelines for the development of word embeddings models \cite{bosc2018auto,faruqui2016problems}, we perform model selection on the dev-set of word relatedness and similarity benchmarks (i.e., SimVerb \cite{gerz2016simverb} and MEN \cite{bruni2014multimodal}). 

\section{Empirical Evaluation}

\begin{table*}[t]
    \small
    \centering
    \begin{tabular}{lccc|cc|cccccc}
        \toprule
         \textbf{Model}  & \textbf{Dim} & \textbf{FT} & \textbf{PT} &  \textbf{SV-d} & \textbf{MEN-d} &  \textbf{SV-t} & \textbf{MEN-t} & \textbf{SL999} & \textbf{SCWS} & \textbf{353} & \textbf{RG}\\ 
       \midrule
       \texttt{Glove} & 300 & yes & no & 12.0 & 54.8 & 7.8 & 57.0 & 19.8 & 46.8 & 44.4 & 57.5\\
       \texttt{Word2Vec} & 300 & yes & no & 35.2 & 62.3 & 36.4  & 59.9 & 34.5 & 54.5 & \textbf{61.9} & 65.7\\
       \midrule
       \texttt{AE} & 300 & yes & no & 34.9 & 42.7 & 32.5 & 42.2 & 35.6 &  50.2 & 41.4 & 64.8\\
       \texttt{CPAE} & 300  & yes & no & 42.8 & 48.5 & 34.8 & 49.2 & 39.5 & 54.3 & 48.7 & 67.1 \\
       \texttt{CPAE-P} & 300  & yes & yes & 44.1 & \textbf{65.1}	& 42.3 & 63.8 & 45.8 & \textbf{60.4} & 61.3 & 72.0\\
       \midrule
       \texttt{bert-base} & 768 & no & yes & 13.5 & 27.8 & 13.3 & 30.6 & 15.1 & 37.8 & 20.0 & 68.1 \\
       \texttt{bert-large} & 1024 & no & yes & 16.1 & 23.4 & 14.4 & 26.8 & 13.4 & 35.7 & 19.8 & 60.7 \\
       \midrule
        \texttt{defsent-bert}  & 768 & yes & yes & 40.0 & 60.2 & 40.0 & 60.0 & 42.0 & 56.8 & 46.6 & 82.4\\
        \texttt{defsent-roberta}  & 768 & yes & yes & 43.0 & 55.0 & 44.0 & 52.6 & 47.7 & 54.3 & 44.9 & 80.6\\
        \midrule
      
        \texttt{distilroberta-v1} & 768 & no & yes & 35.8 & 61.2 & 36.7 & 62.2 & 43.4 & 57.1 & 52.0 & 77.4\\
       \texttt{mpnet-base-v2} & 768 & no & yes & 45.9 & 64.9 & 42.5 & \textbf{67.5} & 49.5 & 58.6 & 56.5 & 81.3\\
        \texttt{sentence-t5-large} & 768 & no & yes & \textbf{49.4} & 63.1 & \underline{\textbf{50.2}} & 66.3 & \underline{\textbf{57.3}} & 56.1 & 51.8 & \underline{\textbf{85.3}}\\
       \midrule
       \textbf{Multi-Relational} \\
       \midrule
       \texttt{Euclidean} & 40 & yes & no & 39.1 & 62.9 & 35.7 & 65.4 & 36.3 & 58.2 & 52.1 & 80.9\\
       \texttt{Euclidean} & 80 & yes & no & 44.1 & 65.6 & 39.5 & 66.2 & 41.2 & 58.4 & 55.8 & 78.0\\
       \texttt{Euclidean} & 200 & yes & no & 47.3 & 67.0 & 41.0 & 67.6 & 43.4 & 60.6 & 55.4 & 78.1\\
       \texttt{Euclidean} & 300 & yes & no & 47.9 & 68.3 & 43.1 & 69.1 & \textbf{44.7} & 61.0 & 54.4 & 79.0\\
       \midrule
       \texttt{Hyperbolic} & 40 & yes & no & 36.7 & 66.2 & 34.3 & 66.4 & 31.8 & 57.7 & 49.9 & 75.5\\
       \texttt{Hyperbolic} & 80 & yes & no & 42.7 & 68.2 & 40.7 & 68.6 & 38.3 & 60.5 & 57.3 & 81.0\\
       \texttt{Hyperbolic} & 200 & yes & no & 48.8 & 71.9 & 44.7 & 73.2 & 40.7 & 62.5 & 62.5 & \textbf{81.6}\\
       \texttt{Hyperbolic} & 300 & yes & no & \underline{\textbf{50.6}} & \underline{\textbf{72.6}} & \textbf{45.4} & \underline{\textbf{74.2}} & 42.3 & \underline{\textbf{63.0}} & \underline{\textbf{63.3}} & 80.5\\
       \bottomrule
        \end{tabular}
    \caption{Results on word similarity and relatedness benchmarks (Spearman's correlation). The column \textbf{FT} indicates whether the model is explicitly fine-tuned on natural language definitions, while \textbf{PT} indicates the adoption of a pre-training phase on external corpora.}
    \label{tab:compare_approaches_overall}
\end{table*}

\begin{table}[t]
    \small
    \centering
    \begin{tabular}{lccccc}
        \toprule
         \textbf{Model} & \textbf{SV} & \textbf{MEN} & \textbf{SL999} & \textbf{353} & \textbf{RG}\\ 
       \midrule
       \texttt{Glove} & 18.9 & - & 32.1 & 62.1 & 75.8\\
       \texttt{Word2Vec} & - & 72.2 & 28.3 & \textbf{68.4} & -\\
       \midrule
       \texttt{Our} & \textbf{45.4} & \textbf{74.2} & \textbf{42.3} & 63.3 & \textbf{80.5}\\
       \bottomrule
        \end{tabular}
    \caption{Comparison with Hyperbolic word embeddings in the literature. The results for Glove and Word2Vec are taken from \cite{tifrea2018poincare} and \cite{leimeister2018skip} considering their best model.}
    \label{tab:compare_Hyperbolic}
\end{table} 

\subsection{Empirical Setup} 

To assess the quality of the word embeddings, we performed an extensive evaluation on word similarity and relatedness benchmarks in English: SimVerb~\cite{gerz2016simverb}, MEN~\cite{bruni2014multimodal}, SimLex-999~\cite{hill2015simlex}, SCWS~\cite{huang2012improving}, WordSim-353~\cite{finkelstein2001placing} and RG-65~\cite{rubenstein1965contextual}, using WordNet \cite{fellbaum2010wordnet} as the main source of definitions\footnote{\url{https://github.com/tombosc/cpae/blob/master/data/dict_wn.json}}. In particular, we leverage the glosses in WordNet to extract the semantic roles via the methodology described in Section \ref{sec:definition_semantic_roles} and train the multi-relational word embeddings. While WordNet also provides a knowledge graph of linguistic relations, our goal is to test methods that are trained and evaluated \emph{exclusively} on natural language definitions and that can more easily generalise to different dictionaries and definitions in a broader setting. 

The multi-relational word embeddings are trained on a total of $\approx 400k$ definitions from which we are able to extract $\approx 2$ million triples. In order to compare Euclidean and Hyperbolic spaces we train two different versions of the model by specialising the objective function accordingly (Equation \ref{eq:general_translational_objective}). We experiment with varying dimensions for both Euclidean and Hyperbolic embeddings (i.e., 40, 80, 200, and 300), training the models for a total of 300 iterations.
In line with previous work \cite{bosc2018auto,faruqui2016problems}, we evaluate the models on downstream benchmarks comparing the predicted similarity between the pair of words to the ground truth via a Spearman’s correlation coefficient.

\subsection{Baselines}
We evaluate a range of word embedding models on the same set of definitions \cite{bosc2018auto}. Specifically, we compare the proposed multi-relational embeddings against different paradigms adopted in previous work and state-of-the-art approaches. Here, we provide a characterisation of the models adopted for evaluation: 

\paragraph{Distributional.} We compare the multi-relational approach against distributional word embeddings \cite{mikolov2013efficient,pennington2014glove}. Both \texttt{Glove} and \texttt{Word2Vec} have the same dimensionality as the multi-relational approach but are not designed to leverage or preserve explicit semantic relations during training.

\paragraph{Autoencoders.} This paradigm employs encoder-decoder architectures to learn word representations from natural language definitions. In particular, we compare our approach to an autoencoder-based model specialised for natural language definitions known as CPAE \cite{bosc2018auto}, which adopts LSTMs paired with a consistency penalty. Differently from our approach, CPAE requires initialisation with pre-trained word vectors to achieve the best results (i.e., CPAE-P). 

\paragraph{Sentence-Transformers.} Finally, we compare our model against Sentence-Transformers \cite{reimers2019sentence}. 
Here, we use Sentence-Transformers to derive embeddings for the target definienda via the encoding of the corresponding definition sentences in the corpus. As the main function of definitions is to describe the meaning of words, semantically similar words tend to possess similar definitions; therefore we expect Sentence-Transformers to organise the latent space in a semantically coherent manner when using definition sentences as a proxy for the word embeddings. We experiment with a diverse set of models ranging from BERT \cite{devlin2019bert} to the current state-of-the-art on semantic similarity benchmarks\footnote{\url{https://www.sbert.net/docs/pretrained_models.html}} \cite{ni2022sentence,song2020mpnet,liu2019roberta} and models trained directly on definition sentences (e.g., Defsent \cite{tsukagoshi2021defsent}). While the evaluated Transformers do not require fine-tuning on the word similarity benchmarks, they are employed after being extensively pre-trained on large corpora and specialised in sentence-level semantic tasks. Moreover, the overall size of the resulting embeddings is significantly larger than the proposed multi-relational approach.

\subsection{Word Embeddings Benchmarks}

In this section, we discuss and analyse the quantitative results obtained on the word similarity and relatedness benchmarks (see Table \ref{tab:compare_approaches_overall}).

Firstly, an internal comparison between Euclidean and Hyperbolic embeddings supports the central hypothesis that Hyperbolic manifolds are particularly suitable for encoding the recursive and hierarchical structure of definitions. As the dimensions of the embeddings increase, the quantitative analysis demonstrates that the Hyperbolic model can achieve the best performance on the majority of the benchmarks.

When compared to the distributional baselines, the multi-relational Hyperbolic embeddings clearly outperform both \texttt{Glove} and \texttt{Word2Vec} trained on the same set of definitions. Similar results can be observed when considering the autoencoder paradigm (apart from CPAE-P on SL999). Since the size of the embeddings produced by the models is comparable (i.e., 300 dimensions), we attribute the observed results to the encoded semantic relations, which might play a crucial role in imposing structural constraints during training.

Finally, the multi-relational model produces embeddings that are competitive with state-of-the-art Transformers. While the Hyperbolic approach can clearly outperform BERT on all the downstream tasks, we observe that Sentence-Transformers become increasingly more competitive when considering larger models that are fine-tuned on semantic similarity tasks and definitions (e.g., \texttt{sentence-t5-large} \cite{ni2022sentence} and \texttt{defsent}  \cite{tsukagoshi2021defsent}). However, it is important to notice that the multi-relational embeddings not only require a small fraction of the Transformers' computational cost -- e.g, T5-large \cite{raffel2020exploring} is pre-trained on the C4 corpus ($\approx$ 750GB) while the multi-relational embeddings are only trained on WordNet glosses ($\approx$ 19MB), a difference of $4$ orders of magnitude -- but are also intrinsically more interpretable thanks to the explicit encoding of the semantic relations (see Section \ref{sec:semantic_control} and \ref{sec:qualitative_analysis}).

\begin{table*}[t]
    \small
    \centering
    \scalebox{1.0}{
    \begin{tabular}{lc|cccc|ccccc}
        \toprule
\multicolumn{1}{c}{\multirow{2}{*}{\textbf{Model}}} & \multicolumn{1}{c}{\multirow{2}{*}{\textbf{Dimension}}} & \multicolumn{2}{c}{\textbf{Mean-Pooling}}              & \multicolumn{2}{c}{\textbf{Multi-Relational}} & \multicolumn{2}{c}{\textbf{Differentia Quality}}  & \multicolumn{2}{c}{\textbf{Supertype}}\\ 
\cmidrule(l){3-10} 
\multicolumn{1}{c}{} & \multicolumn{1}{c}{} & \textbf{SV}  & \textbf{MEN} & \textbf{SV}  & \textbf{MEN} & \textbf{SV}  & \textbf{MEN} & \textbf{SV}  & \textbf{MEN}\\ \midrule
       \texttt{Euclidean} & 40 &\textbf{17.6} & \textbf{20.6} & 23.7 (+6.1)& \textbf{31.7} (+11.1) & 22.6 & \textbf{26.0} & 17.2 & 19.2\\
       \texttt{Euclidean} & 80 & 15.9 & 18.1 & \textbf{24.6} (+8.7) & 29.4 (+11.3) & 23.4 & 23.3 & 18.4 & 18.8\\
       \texttt{Euclidean} & 200 & 14.5 & 18.4 & 23.7 \textbf{(+9.2)} & 30.7 \textbf{(+12.3)} & \textbf{24.1} & 22.2 & 18.7 & 19.1\\
       \texttt{Euclidean} &  300 & 15.1 & 18.8 & 24.3 \textbf{(+9.2)} & 30.3 (+11.5) & 23.8 & 22.7 & \underline{\textbf{19.3}} & \textbf{20.3}\\
       \midrule
       \texttt{Hyperbolic}  & 40 & 15.9 & 22.8 & 25.4 (+9.5) & 35.2 (+12.4)  & 22.7 & 25.5 & 14.0 & 20.2\\
       \texttt{Hyperbolic}  & 80 & 17.9 & 25.1 & 27.7 \underline{\textbf{(+9.8)}} & 37.8 (+12.7) & 25.9 & \underline{\textbf{26.6}} & 15.4 & 20.1\\
        \texttt{Hyperbolic}  & 200 & 19.2 & 24.9 & 28.4 (+9.2)& 38.2 (+13.3) & 27.9 & 25.5 & 17.3 & \underline{\textbf{21.3}}\\
       \texttt{Hyperbolic}  & 300 & \underline{\textbf{19.6}} & \underline{\textbf{25.1}}  & \underline{\textbf{28.6}} (+9.0) & \underline{\textbf{39.7 (+14.6)}}  &\underline{\textbf{28.5}} & 26.0 & \textbf{18.1} & 20.4\\
       \bottomrule
        \end{tabular}}
    \caption{Results on the one-shot approximation of out-of-vocabulary word embeddings. The numbers in the table represent the Spearman correlation computed over the out-of-vocabulary set after the approximation. (Left) impact of the multi-relational embeddings on the one-shot encoding of out-of-vocabulary words. (Right) ablations using the two most common semantic roles for one-shot approximation. The results demonstrate the superior capacity of the multi-relational Hyperbolic embeddings to capture the global semantic structure of definitions.}
    \label{tab:compare_one_shot_approaches}
\end{table*}

\subsection{Hyperbolic Word Embeddings}
In addition to the previous analysis, we performed a comparison with existing Hyperbolic word embeddings in the literature (Table \ref{tab:compare_Hyperbolic}). In particular, we compare the proposed multi-relational model with Poincare Glove \cite{tifrea2018poincare} and Hyperbolic Word2Vec \cite{leimeister2018skip}. The results show that our approach can outperform both models on the majority of the benchmarks, remarking the impact of the multi-relational approach and definitional model on the quality of the representation.

\subsection{Multi-Relational Representation}
\label{sec:semantic_control}

To contrast the capacity of different geometric spaces to learn multi-relational representations, we design an additional experiment that tests the ability to encode out-of-vocabulary definienda (i.e., words never seen during training). In particular, we aim to quantitatively measure the precision in encoding the semantic relations by approximating new word embeddings in one-shot, and use it as a proxy for assessing the structural organisation of Euclidean and Hyperbolic spaces. Our hypothesis is that a vector space organised according to the multi-relational structure induced by the definitions should allow for a more precise approximation of out-of-vocabulary word embeddings via relation-specific transformations.

In order to perform this experiment, we adopt the dev-set of SimVerb \cite{gerz2016simverb} and MEN \cite{bruni2014multimodal}, removing all the triples from our training set that contain a subject or an object word occurring in the benchmarks. Subsequently, we employ the pruned training set to re-train the models.
After training, we derive the embeddings of the out-of-vocabulary words via geometric transformations applied to the in-vocabulary words.
Specifically, given a target word (e.g., "dog") and its definition (e.g., "a domesticated carnivorous mammal that typically has a long snout") we jointly use the in-vocabulary definiens and their semantic relations (e.g., ["carnivorous", "supertype"], ["snout", "differentia-quality"]) to approximate a new word embedding e() for the definiendum via mean pooling and translation (i.e., e("dog") = mean(e("carnivorous"), ("snout")) + mean(e("supertype"), e("differentia-quality"))) and compare against a mean pooling baseline that does not have access to the semantic relations (i.e. e("dog") = mean(e("carnivorous"), ("snout"))).

The results reported in Table \ref{tab:compare_one_shot_approaches} demonstrate the impact of the multi-relational framework, also confirming the property of the Hyperbolic embeddings in better encoding the global semantic structure of natural language definitions.

\begin{table*}[t]
    \tiny
    \centering
    \begin{tabular}{llp{5.6cm}p{5.6cm}}
        \toprule
        \textbf{Category} & \textbf{Word Pair} &  \textbf{Euclidean} & \textbf{Hyperbolic} \\ 
         \midrule
         Concrete concepts &
         car - bycicle & 
         bicycle,
         car,
         pedal\_driven,
         motorcycle,
         banked,
         multiplying,
         swiveling,
         four\_wheel,
         rented,
         no\_parking & 
        railcar,
        bicycle,
        car,
        pedal\_driven,
        driving\_axle,
        motorized\_wheelchair,
        tricycle,
        bike,
        banked,
        live\_axle
        \\
         \midrule
         Gender, Role &
         man - woman & 
         woman,
         man,
         procreation,
         men,
         non\_jewish,
         three\_cornered,
         middle\_aged,
         bodice,
         boskop,
         soloensis & 
        adulterer,
        boyfriend,
        ex-boyfriend,
        adult\_female,
        manful,
        cuckold,
        virile,
        stateswoman,
        womanlike,
        wardress
         \\\midrule
        Animal Hybrids &

        horse - donkey &
        donkey,
        horse,
        burro,
        hock\_joint,
        neighing,
        dog\_sized,
        tapirs,
        feathered\_legged,
        racehorse,
        gasterophilidae & 
        burro,
        cow\_pony,
        unbridle,
        hackney,
        unbridled,
        equitation,
        sidesaddle,
        palfrey,
        roughrider,
        trotter
        \\\midrule          
        Process, Time &
        birth - death & 
        death,
        birth,
        lifetime,
        childless,
        childhood,
        adityas,
        parturition,
        condemned,
        carta,
        liveborn& 
        lifespan,
        life-time,
        firstborn,
        multiparous,
        full\_term,
        teens,
        nonpregnant,
        childless,
        widowhood,
        gestational\\
         \midrule

        Location &

        sea - land & 
        land,
        sea,
        enderby,
        weddell,
        arafura,
        littoral,
        tyrrhenian,
        andaman,
        maud,
        toads& 
        tellurian,
        litoral,
        seabed,
        high\_sea,
        body\_of\_water,
        littoral\_zone,
        international\_waters,
        benthic\_division,
        naval\_forces,
        lake\_michigan\\
       \bottomrule
        \end{tabular}
        \newline
        \newline
    \begin{tabular}{lp{7cm}p{7cm}}
        \toprule
         \textbf{$w_s$} &  \textbf{No Transformation:} $-d_\mathbb{B}(\textbf{h}_s, \textbf{h}_o)^2$ & \textbf{Relation-Adjusted} ($r$ = \texttt{supertype})\textbf{:} $-d_\mathbb{B}({R}\otimes\textbf{h}_s, \textbf{h}_o \oplus \textbf{r})^2$ \\ 
         \midrule
         dog & dog, heavy\_coated, smooth\_coated, malamute, canidae, wolves, light\_footed, long-established, whippet, greyhound
         & huntsman, hunting\_dog, sledge\_dog, coondog, sled\_dog, working\_dog, russian\_wolfhound, guard\_dog, tibetan\_mastiff, housedog 
         \\
         \midrule
         car & car, railcar, telpherage, telferage, subcompact, cable\_car, car\_transporter, re\_start, auto, railroad\_car, driving\_axle
         & railcar, marksman, subcompact, smoking\_carriage, handcar, electric\_automobile, limousine, taxicab, freight\_car
         , slip\_coach 
         \\
         \midrule
         star & star, armillary\_sphere, charles's\_wain, starlight, altair, drummer, northern\_cross, photosphere, sterope, rigel
         & rigel, betelgeuse, film\_star, movie\_star, television\_star, tv\_star, starlight, supergiant, photosphere, starlet
         \\
         \midrule
         king & louis\_i, sultan, sir\_gawain. uriah, camelot, dethrone, poitiers, excalibur, empress, divorcee 
         & chessman, gustavus\_vi, grandchild, alfred\_the\_great, jr, rajah, knights, louis\_the\_far, egbert, plantagenet, st.\_olav
         \\
       \bottomrule
        \end{tabular}
       \caption{(Top) qualitative results for the latent space traversal, with midpoint nearest neighbours listed in descending order. (Bottom) nearest neighbours of seed words before and after applying a supertype-adjusted transformation. }
        \label{tab:qualitative_analysis}
\end{table*}

\section{Qualitative Analysis}
\label{sec:qualitative_analysis}
In addition to the qualitative evaluation, we perform a qualitative analysis of the embeddings. 
This is performed in two different ways: traversal of the latent space and relation-adjusted transformations.

\subsection{Latent Space Traversal}

We perform traversal experiments to visualise the organisation of the latent space. This is done by sampling points at fixed intervals along the arc (i.e., geodesic) connecting the embeddings of a pair of predefined words (seeds), i.e., by interpolating along the shortest path between two embeddings. The choice of word pairs was done according to a group of semantic categories for which intermediate concepts can be understood to be semantically in between the pair. For example: $(car, bicycle) \rightarrow motorcycle$. Considering the latent space structure that should result from the proposed approach, we expect the traversal process to capture such intermediate concepts, while generalising the concepts towards the midpoint of the arc. In a latent space organised according to the semantic structure and concept hierarchy of definitions, in fact, we expect the midpoint to be close to concepts relating to both seed words.

The categories, sampled words and results for the midpoint of the arcs can be found in Table~\ref{tab:qualitative_analysis} (top).
From the traversal analysis, we can observe that the intermediate concepts are indeed captured for all the categories, with a noticeable degree of generalisation in the Hyperbolic models. This indicates the consistent interpretable nature of the navigation for the latent space, and enables more robust semantic control, setting the desired embedded concept in terms of a symbolic conjunction of its vicinity. We can also observe that, the space between the pair of embeddings is populated mostly by concepts \textit{related} to both entities of the pair in the Euclidean models, while being populated by concepts \textit{relating} both entities in the Hyperbolic models. 

\subsection{Relation-Adjusted Transformations}
We analyse the organisation of the latent space before and after the application of a translational operation.
As discussed in Section~\ref{sec:dictionary_definitions}, such operation should transform the embedding space according to the corresponding semantic role. 
For example, the operation $\phi_\mathbb{B}(dog, supertype, w_o)$ should cluster the space around the taxonomical branch related to \emph{``dog''}. It is important to notice that this operation does not correspond to link prediction as we are not considering the scalar biases $b_s, b_o$. The goal here is to disentangle the impact of the semantic transformations on the latent space. 
We consider the \texttt{supertype} role for this analysis as it induces a global hierarchical structure that is easily inspectable. The results can be found in Table~\ref{tab:qualitative_analysis} (bottom). We observe that the transformation leads to a projection locus near all the closely defined terms (the types of dogs or stars), abstracting the subject words in terms of their conceptual extension (things that are dogs / stars). This displays a particular way of generalisation that is likely related to the arrangement of the roles and how they connect the concepts.

\section{Related Work}
Considering the basic characteristics of natural language definitions here discussed, efforts to leverage dictionary definitions for distributional models were proposed as a more efficient alternative to the large unlabeled corpora, following the rising popularity of the latter \cite{tsukagoshi2021defsent,hill2016learning, tissier2017dict2vec, bosc2018auto}. 
Simultaneously, efforts to improve compositionality~\cite{chen2015improving, scheepers2018improving} and interpretability~\cite{carvalho2017building, silva2019exploring} of word representations led to different approaches towards the incorporation of definition resources to language modelling, with the idea of modelling definitions becoming an established task~\cite{noraset2017definition}.

More recently, research focus has shifted towards the fine-tuning of large language models and contextual embeddings for definition generation and classification~\cite{gadetsky2018conditional, bosc2018auto, loureiro2019language, mickus2022semeval}, with interest in the structural properties of definitions also gaining attention \cite{shu2020drg2vec, wang2022hg2vec}.

Finally, research on Hyperbolic representation spaces has provided evidence of improvements in capturing hierarchical linguistic features, over traditional (Euclidean) ones \cite{balazevic2019multi,nickel2017poincare,tifrea2018poincare, zhao2020manifold}.
This work builds upon the aforementioned developments, and proposes a novel approach to the incorporation of structural information extracted from natural language definitions by means of a translational objective guided by explicit semantic roles \cite{silva2016categorization}, combined with a Hyperbolic representation able to embed multi-relational structures.   

\section{Conclusion}

This paper explored the semantic structure of \emph{definitions} as a means to support novel learning paradigms able to preserve semantic interpretability and control. We proposed a \emph{multi-relational} framework that can explicitly map terms and their corresponding \emph{semantic relations} into a vector space. By automatically extracting the relations from external dictionaries, and specialising the framework in \emph{Hyperbolic space}, we demonstrated that it is possible to capture the hierarchical and multi-relational structure induced by dictionary definitions while preserving, at the same time, the explicit mapping required for controllable semantic navigation. 

\section{Limitations}
While the study here presented supports its findings with all the evidence compiled to the best of our knowledge, there are factors that limit the scope of the current state of the work, from which we understand as the most important: 
\begin{enumerate}
    \item The automatic semantic role labeling process is not 100\% accurate, and thus is a limiting factor in analysing the impact of this information on the models. While we do not explore DSRLs with varying accuracy, future work can explicitly investigate the impact of the automatic annotation on the robustness of the multi-relational embeddings.
    \item The embeddings obtained in this work are contextualizable (by means of a relation-adjusted transformation), but are not \textit{contextualized}, i.e., they are not dependent on surrounding text. Therefore, they are not comparable on tasks dependant on contextualised embeddings.
    \item The current version of the embeddings coalesces all senses of a definiendum into a single representation. This is a general limitation of models learning embeddings from dictionaries. Fixing this limitation is possible in future work, but it will require the non-trivial ability to disambiguate the terms appearing in the definitions (i.e., definiens).
    \item The multi-relational embeddings presented in the paper were initialised from scratch in order to test their efficiency in capturing the semantic structure of dictionary definitions. Therefore, there is an open question regarding the possible benefits of initialising the models with pre-trained distributional embeddings such as \texttt{Word2Vec} and \texttt{Glove}. 
\end{enumerate}

\section*{Acknowledgements}
This work was partially funded by the Swiss National Science Foundation (SNSF) project NeuMath (\href{https://data.snf.ch/grants/grant/204617}{200021\_204617}), by the EPSRC grant EP/T026995/1 entitled “EnnCore: End-to-End Conceptual Guarding of Neural Architectures” under Security for all in an AI enabled society, by the CRUK National Biomarker Centre, and supported by the Manchester Experimental Cancer Medicine Centre and the NIHR Manchester Biomedical Research Centre.

\bibliography{anthology,custom}

\begin{thebibliography}{46}
\expandafter\ifx\csname natexlab\endcsname\relax\def\natexlab#1{#1}\fi

\bibitem[{Balazevic et~al.(2019)Balazevic, Allen, and
  Hospedales}]{balazevic2019multi}
Ivana Balazevic, Carl Allen, and Timothy Hospedales. 2019.
\newblock Multi-relational poincar{\'e} graph embeddings.
\newblock \emph{Advances in Neural Information Processing Systems}, 32.

\bibitem[{Bordes et~al.(2013)Bordes, Usunier, Garcia-Duran, Weston, and
  Yakhnenko}]{bordes2013translating}
Antoine Bordes, Nicolas Usunier, Alberto Garcia-Duran, Jason Weston, and Oksana
  Yakhnenko. 2013.
\newblock Translating embeddings for modeling multi-relational data.
\newblock \emph{Advances in neural information processing systems}, 26.

\bibitem[{Bosc and Vincent(2018)}]{bosc2018auto}
Tom Bosc and Pascal Vincent. 2018.
\newblock Auto-encoding dictionary definitions into consistent word embeddings.
\newblock In \emph{Proceedings of the 2018 Conference on Empirical Methods in
  Natural Language Processing}, pages 1522--1532.

\bibitem[{Bruni et~al.(2014)Bruni, Tran, and Baroni}]{bruni2014multimodal}
Elia Bruni, Nam-Khanh Tran, and Marco Baroni. 2014.
\newblock Multimodal distributional semantics.
\newblock \emph{Journal of artificial intelligence research}, 49:1--47.

\bibitem[{Carvalho et~al.(2023)Carvalho, Mercatali, Zhang, and
  Freitas}]{carvalho2022learning}
Danilo~S Carvalho, Giangiacomo Mercatali, Yingji Zhang, and Andre Freitas.
  2023.
\newblock Learning disentangled representations for natural language
  definitions.
\newblock \emph{Findings of the European chapter of Association for
  Computational Linguistics (Findings of EACL)}.

\bibitem[{Chen et~al.(2015)Chen, Xu, He, and Wang}]{chen2015improving}
Tao Chen, Ruifeng Xu, Yulan He, and Xuan Wang. 2015.
\newblock Improving distributed representation of word sense via wordnet gloss
  composition and context clustering.
\newblock In \emph{Proceedings of the 53rd Annual Meeting of the Association
  for Computational Linguistics and the 7th International Joint Conference on
  Natural Language Processing (Volume 2: Short Papers)}, pages 15--20.

\bibitem[{de~Carvalho and Le~Nguyen(2017)}]{carvalho2017building}
Danilo~Silva de~Carvalho and Minh Le~Nguyen. 2017.
\newblock Building lexical vector representations from concept definitions.
\newblock In \emph{Proceedings of the 15th Conference of the European Chapter
  of the Association for Computational Linguistics: Volume 1, Long Papers},
  pages 905--915.

\bibitem[{Devlin et~al.(2019)Devlin, Chang, Lee, and
  Toutanova}]{devlin2019bert}
Jacob Devlin, Ming-Wei Chang, Kenton Lee, and Kristina Toutanova. 2019.
\newblock Bert: Pre-training of deep bidirectional transformers for language
  understanding.
\newblock In \emph{Proceedings of the 2019 Conference of the North American
  Chapter of the Association for Computational Linguistics: Human Language
  Technologies, Volume 1 (Long and Short Papers)}, pages 4171--4186.

\bibitem[{Faruqui et~al.(2016)Faruqui, Tsvetkov, Rastogi, and
  Dyer}]{faruqui2016problems}
Manaal Faruqui, Yulia Tsvetkov, Pushpendre Rastogi, and Chris Dyer. 2016.
\newblock Problems with evaluation of word embeddings using word similarity
  tasks.
\newblock In \emph{Proceedings of the 1st Workshop on Evaluating Vector-Space
  Representations for NLP}, pages 30--35.

\bibitem[{Fellbaum(2010)}]{fellbaum2010wordnet}
Christiane Fellbaum. 2010.
\newblock Wordnet.
\newblock In \emph{Theory and applications of ontology: computer applications},
  pages 231--243. Springer.

\bibitem[{Feng et~al.(2016)Feng, Huang, Wang, Zhou, Hao, and
  Zhu}]{feng2016knowledge}
Jun Feng, Minlie Huang, Mingdong Wang, Mantong Zhou, Yu~Hao, and Xiaoyan Zhu.
  2016.
\newblock Knowledge graph embedding by flexible translation.
\newblock In \emph{Fifteenth International Conference on the Principles of
  Knowledge Representation and Reasoning}.

\bibitem[{Finkelstein et~al.(2001)Finkelstein, Gabrilovich, Matias, Rivlin,
  Solan, Wolfman, and Ruppin}]{finkelstein2001placing}
Lev Finkelstein, Evgeniy Gabrilovich, Yossi Matias, Ehud Rivlin, Zach Solan,
  Gadi Wolfman, and Eytan Ruppin. 2001.
\newblock Placing search in context: The concept revisited.
\newblock In \emph{Proceedings of the 10th international conference on World
  Wide Web}, pages 406--414.

\bibitem[{Gadetsky et~al.(2018)Gadetsky, Yakubovskiy, and
  Vetrov}]{gadetsky2018conditional}
Artyom Gadetsky, Ilya Yakubovskiy, and Dmitry Vetrov. 2018.
\newblock Conditional generators of words definitions.
\newblock In \emph{Proceedings of the 56th Annual Meeting of the Association
  for Computational Linguistics (Volume 2: Short Papers)}, pages 266--271.

\bibitem[{Ganea et~al.(2018)Ganea, B{\'e}cigneul, and
  Hofmann}]{ganea2018Hyperbolic}
Octavian Ganea, Gary B{\'e}cigneul, and Thomas Hofmann. 2018.
\newblock Hyperbolic neural networks.
\newblock \emph{Advances in neural information processing systems}, 31.

\bibitem[{Gerz et~al.(2016)Gerz, Vuli{\'c}, Hill, Reichart, and
  Korhonen}]{gerz2016simverb}
Daniela Gerz, Ivan Vuli{\'c}, Felix Hill, Roi Reichart, and Anna Korhonen.
  2016.
\newblock Simverb-3500: A large-scale evaluation set of verb similarity.
\newblock In \emph{Proceedings of the 2016 Conference on Empirical Methods in
  Natural Language Processing}, pages 2173--2182.

\bibitem[{Hill et~al.(2016)Hill, Cho, Korhonen, and Bengio}]{hill2016learning}
Felix Hill, Kyunghyun Cho, Anna Korhonen, and Yoshua Bengio. 2016.
\newblock Learning to understand phrases by embedding the dictionary.
\newblock \emph{Transactions of the Association for Computational Linguistics},
  4:17--30.

\bibitem[{Hill et~al.(2015)Hill, Reichart, and Korhonen}]{hill2015simlex}
Felix Hill, Roi Reichart, and Anna Korhonen. 2015.
\newblock Simlex-999: Evaluating semantic models with (genuine) similarity
  estimation.
\newblock \emph{Computational Linguistics}, 41(4):665--695.

\bibitem[{Huang et~al.(2012)Huang, Socher, Manning, and
  Ng}]{huang2012improving}
Eric~H Huang, Richard Socher, Christopher~D Manning, and Andrew~Y Ng. 2012.
\newblock Improving word representations via global context and multiple word
  prototypes.
\newblock In \emph{Proceedings of the 50th Annual Meeting of the Association
  for Computational Linguistics (Volume 1: Long Papers)}, pages 873--882.

\bibitem[{Leimeister and Wilson(2018)}]{leimeister2018skip}
Matthias Leimeister and Benjamin~J Wilson. 2018.
\newblock Skip-gram word embeddings in hyperbolic space.
\newblock \emph{arXiv preprint arXiv:1809.01498}.

\bibitem[{Liu et~al.(2019)Liu, Ott, Goyal, Du, Joshi, Chen, Levy, Lewis,
  Zettlemoyer, and Stoyanov}]{liu2019roberta}
Yinhan Liu, Myle Ott, Naman Goyal, Jingfei Du, Mandar Joshi, Danqi Chen, Omer
  Levy, Mike Lewis, Luke Zettlemoyer, and Veselin Stoyanov. 2019.
\newblock Roberta: A robustly optimized bert pretraining approach.
\newblock \emph{arXiv preprint arXiv:1907.11692}.

\bibitem[{Loureiro and Jorge(2019)}]{loureiro2019language}
Daniel Loureiro and Alipio Jorge. 2019.
\newblock Language modelling makes sense: Propagating representations through
  wordnet for full-coverage word sense disambiguation.
\newblock In \emph{Proceedings of the 57th Annual Meeting of the Association
  for Computational Linguistics}, pages 5682--5691.

\bibitem[{Mickus et~al.(2022)Mickus, van Deemter, Constant, and
  Paperno}]{mickus2022semeval}
Timothee Mickus, Kees van Deemter, Mathieu Constant, and Denis Paperno. 2022.
\newblock Semeval-2022 task 1: Codwoe--comparing dictionaries and word
  embeddings.
\newblock \emph{arXiv preprint arXiv:2205.13858}.

\bibitem[{Mikolov et~al.(2013)Mikolov, Chen, Corrado, and
  Dean}]{mikolov2013efficient}
Tomas Mikolov, Kai Chen, Greg Corrado, and Jeffrey Dean. 2013.
\newblock Efficient estimation of word representations in vector space.
\newblock \emph{arXiv preprint arXiv:1301.3781}.

\bibitem[{Miller(1995)}]{miller1995wordnet}
George~A Miller. 1995.
\newblock Wordnet: a lexical database for english.
\newblock \emph{Communications of the ACM}, 38(11):39--41.

\bibitem[{Ni et~al.(2022)Ni, Abrego, Constant, Ma, Hall, Cer, and
  Yang}]{ni2022sentence}
Jianmo Ni, Gustavo~Hernandez Abrego, Noah Constant, Ji~Ma, Keith Hall, Daniel
  Cer, and Yinfei Yang. 2022.
\newblock Sentence-t5: Scalable sentence encoders from pre-trained text-to-text
  models.
\newblock In \emph{Findings of the Association for Computational Linguistics:
  ACL 2022}, pages 1864--1874.

\bibitem[{Nickel and Kiela(2017)}]{nickel2017poincare}
Maximillian Nickel and Douwe Kiela. 2017.
\newblock Poincar{\'e} embeddings for learning hierarchical representations.
\newblock \emph{Advances in neural information processing systems}, 30.

\bibitem[{Noraset et~al.(2017)Noraset, Liang, Birnbaum, and
  Downey}]{noraset2017definition}
Thanapon Noraset, Chen Liang, Larry Birnbaum, and Doug Downey. 2017.
\newblock Definition modeling: Learning to define word embeddings in natural
  language.
\newblock In \emph{Thirty-First AAAI Conference on Artificial Intelligence}.

\bibitem[{Pennington et~al.(2014)Pennington, Socher, and
  Manning}]{pennington2014glove}
Jeffrey Pennington, Richard Socher, and Christopher~D Manning. 2014.
\newblock Glove: Global vectors for word representation.
\newblock In \emph{Proceedings of the 2014 conference on empirical methods in
  natural language processing (EMNLP)}, pages 1532--1543.

\bibitem[{Raffel et~al.(2020)Raffel, Shazeer, Roberts, Lee, Narang, Matena,
  Zhou, Li, and Liu}]{raffel2020exploring}
Colin Raffel, Noam Shazeer, Adam Roberts, Katherine Lee, Sharan Narang, Michael
  Matena, Yanqi Zhou, Wei Li, and Peter~J Liu. 2020.
\newblock Exploring the limits of transfer learning with a unified text-to-text
  transformer.
\newblock \emph{Journal of Machine Learning Research}, 21:1--67.

\bibitem[{Reimers and Gurevych(2019)}]{reimers2019sentence}
Nils Reimers and Iryna Gurevych. 2019.
\newblock Sentence-bert: Sentence embeddings using siamese bert-networks.
\newblock In \emph{Proceedings of the 2019 Conference on Empirical Methods in
  Natural Language Processing and the 9th International Joint Conference on
  Natural Language Processing (EMNLP-IJCNLP)}, pages 3982--3992.

\bibitem[{Rubenstein and Goodenough(1965)}]{rubenstein1965contextual}
Herbert Rubenstein and John~B Goodenough. 1965.
\newblock Contextual correlates of synonymy.
\newblock \emph{Communications of the ACM}, 8(10):627--633.

\bibitem[{Sanh et~al.(2019)Sanh, Debut, Chaumond, and
  Wolf}]{sanh2019distilbert}
Victor Sanh, Lysandre Debut, Julien Chaumond, and Thomas Wolf. 2019.
\newblock Distilbert, a distilled version of bert: smaller, faster, cheaper and
  lighter.
\newblock \emph{arXiv preprint arXiv:1910.01108}.

\bibitem[{Scheepers et~al.(2018)Scheepers, Kanoulas, and
  Gavves}]{scheepers2018improving}
Thijs Scheepers, Evangelos Kanoulas, and Efstratios Gavves. 2018.
\newblock Improving word embedding compositionality using lexicographic
  definitions.
\newblock In \emph{Proceedings of the 2018 World Wide Web Conference}, pages
  1083--1093.

\bibitem[{Shu et~al.(2020)Shu, Yu, Zhang, and Liu}]{shu2020drg2vec}
Xiaobo Shu, Bowen Yu, Zhenyu Zhang, and Tingwen Liu. 2020.
\newblock Drg2vec: Learning word representations from definition relational
  graph.
\newblock In \emph{2020 International Joint Conference on Neural Networks
  (IJCNN)}, pages 1--9. IEEE.

\bibitem[{Silva et~al.(2018{\natexlab{a}})Silva, Freitas, and
  Handschuh}]{silva2018building}
Vivian Silva, Andr{\'e} Freitas, and Siegfried Handschuh. 2018{\natexlab{a}}.
\newblock Building a knowledge graph from natural language definitions for
  interpretable text entailment recognition.
\newblock In \emph{Proceedings of the Eleventh International Conference on
  Language Resources and Evaluation (LREC 2018)}.

\bibitem[{Silva et~al.(2016)Silva, Handschuh, and
  Freitas}]{silva2016categorization}
Vivian Silva, Siegfried Handschuh, and Andr{\'e} Freitas. 2016.
\newblock Categorization of semantic roles for dictionary definitions.
\newblock In \emph{Proceedings of the 5th Workshop on Cognitive Aspects of the
  Lexicon (CogALex-V)}, pages 176--184.

\bibitem[{Silva et~al.(2019)Silva, Freitas, and Handschuh}]{silva2019exploring}
Vivian~S Silva, Andr{\'e} Freitas, and Siegfried Handschuh. 2019.
\newblock Exploring knowledge graphs in an interpretable composite approach for
  text entailment.
\newblock In \emph{Proceedings of the AAAI Conference on Artificial
  Intelligence}, volume~33, pages 7023--7030.

\bibitem[{Silva et~al.(2018{\natexlab{b}})Silva, Handschuh, and
  Freitas}]{silva2018recognizing}
Vivian~S Silva, Siegfried Handschuh, and Andr{\'e} Freitas. 2018{\natexlab{b}}.
\newblock Recognizing and justifying text entailment through distributional
  navigation on definition graphs.
\newblock In \emph{Thirty-Second AAAI Conference on Artificial Intelligence}.

\bibitem[{Song et~al.(2020)Song, Tan, Qin, Lu, and Liu}]{song2020mpnet}
Kaitao Song, Xu~Tan, Tao Qin, Jianfeng Lu, and Tie-Yan Liu. 2020.
\newblock Mpnet: Masked and permuted pre-training for language understanding.
\newblock \emph{Advances in Neural Information Processing Systems},
  33:16857--16867.

\bibitem[{Tifrea et~al.(2018)Tifrea, Becigneul, and Ganea}]{tifrea2018poincare}
Alexandru Tifrea, Gary Becigneul, and Octavian-Eugen Ganea. 2018.
\newblock Poincare glove: Hyperbolic word embeddings.
\newblock In \emph{International Conference on Learning Representations}.

\bibitem[{Tissier et~al.(2017)Tissier, Gravier, and
  Habrard}]{tissier2017dict2vec}
Julien Tissier, Christophe Gravier, and Amaury Habrard. 2017.
\newblock Dict2vec: Learning word embeddings using lexical dictionaries.
\newblock In \emph{Proceedings of the 2017 Conference on Empirical Methods in
  Natural Language Processing}, pages 254--263.

\bibitem[{Tsukagoshi et~al.(2021)Tsukagoshi, Sasano, and
  Takeda}]{tsukagoshi2021defsent}
Hayato Tsukagoshi, Ryohei Sasano, and Koichi Takeda. 2021.
\newblock Defsent: Sentence embeddings using definition sentences.
\newblock In \emph{Proceedings of the 59th Annual Meeting of the Association
  for Computational Linguistics and the 11th International Joint Conference on
  Natural Language Processing (Volume 2: Short Papers)}, pages 411--418.

\bibitem[{Ungar(2001)}]{ungar2001Hyperbolic}
Abraham~A Ungar. 2001.
\newblock Hyperbolic trigonometry and its application in the poincar{\'e} ball
  model of hyperbolic geometry.
\newblock \emph{Computers \& Mathematics with Applications}, 41(1-2):135--147.

\bibitem[{Wang and Zaki(2022)}]{wang2022hg2vec}
Qitong Wang and Mohammed~J Zaki. 2022.
\newblock Hg2vec: Improved word embeddings from dictionary and thesaurus based
  heterogeneous graph.
\newblock In \emph{Proceedings of the 29th International Conference on
  Computational Linguistics}, pages 3154--3163.

\bibitem[{Zesch et~al.(2008)Zesch, M{\"u}ller, and Gurevych}]{zesch2008using}
Torsten Zesch, Christof M{\"u}ller, and Iryna Gurevych. 2008.
\newblock Using wiktionary for computing semantic relatedness.
\newblock In \emph{AAAI}, volume~8, pages 861--866.

\bibitem[{Zhao et~al.(2020)Zhao, Zhou, Li, and Chen}]{zhao2020manifold}
Wenyu Zhao, Dong Zhou, Lin Li, and Jinjun Chen. 2020.
\newblock Manifold learning-based word representation refinement incorporating
  global and local information.
\newblock In \emph{Proceedings of the 28th International Conference on
  Computational Linguistics}, pages 3401--3412.

\end{thebibliography}
\bibliographystyle{acl_natbib}

\appendix

\section{Multi-Relational Embeddings}
\label{sec:AppendixA}
The multi-relational embeddings are trained on a total of $\approx 400k$ definitions from which we are able to extract $\approx 2$ million triples. We experiment with varying dimensions for both Euclidean and Hyperbolic embeddings (i.e., 40, 80, 200, and 300), training the models for a total of 300 iterations with batch size 128 and learning rate 50 
 on 16GB Nvidia Tesla V100 GPU. 
The multi-relational models are optimised via a Bernoulli negative log-likelihood loss:
\begin{equation}
\small
    \begin{aligned}
        L(y, p) = -1 \frac{1}{N}\sum_{i=1}^N (y^{(i)}log(p^{(i)}) \\ + (1-y^{(i)})log(1-p^{(i)}))
    \end{aligned}
\end{equation}
where $p^{(i)}$ represents the predictions made by the model and $y^{(i)}$ represents the actual label.


\end{document}